\documentclass[conference]{IEEEtran}
\IEEEoverridecommandlockouts

\usepackage[T1]{fontenc}
\usepackage{cite}
\usepackage[dvipsnames]{xcolor}
\usepackage{graphicx} 
\usepackage{algorithm2e}
\usepackage{tikz}
\usepackage{hyperref}
\usepackage{booktabs}
\usepackage{comment}
\usepackage{float}
\usepackage{tabularx}
\usepackage{makecell}
\usepackage{pifont}
\usepackage{bm} 
\usepackage{amsmath,amssymb,amsfonts}
\usepackage{algorithmic}
\usepackage{subcaption}
\usepackage{textcomp}
\definecolor{fuchsia}{rgb}{155, 0, 255}  
\definecolor{1}{RGB}{255,0,136}
\definecolor{2}{RGB}{255,255,80}
\definecolor{3}{RGB}{255,119,0}
\definecolor{4}{RGB}{0,255,221}
\definecolor{5}{RGB}{187,0,255}
\definecolor{6}{RGB}{0,170,255}
\definecolor{7}{RGB}{0,255,136}
\definecolor{8}{RGB}{102,0,255}
\definecolor{9}{RGB}{0,102,255}
\definecolor{10}{RGB}{0,255,0}
\definecolor{11}{RGB}{255,0,0}
\definecolor{12}{RGB}{0,255,53}
\definecolor{13}{RGB}{61,255,80}
\definecolor{14}{RGB}{200,240,55}
\definecolor{15}{RGB}{240,121,50}
\usepackage{pgfplots}
\usepackage{fancyhdr}
\fancypagestyle{firstpage}{
  \fancyhf{} 
  \fancyhead[L]{\scriptsize The 3rd IEEE International Conference on Federated Learning Technologies and Applications (FLTA25)}

}

\def\BibTeX{{\rm B\kern-.05em{\sc i\kern-.025em b}\kern-.08em
    T\kern-.1667em\lower.7ex\hbox{E}\kern-.125emX}}

\begin{document}

\title{FedQuad: Federated Stochastic Quadruplet Learning to Mitigate Data Heterogeneity}

\author{\"Ozg\"u  G\"oksu\\
School of Computing Science\\
University of Glasgow\\
{\tt\small 2718886G@student.gla.ac.uk}
\and
Nicolas Pugeault\\
School of Computing Science\\
University of Glasgow\\
{\tt\small nicolas.pugeault@glasgow.ac.uk}
}

\maketitle
\thispagestyle{firstpage} 
\begin{abstract}
Federated Learning (FL) provides decentralised model training, which effectively tackles problems such as distributed data and privacy preservation. However, the generalisation of global models frequently faces challenges from data heterogeneity among clients. This challenge becomes even more pronounced when datasets are limited in size and class imbalance. To address data heterogeneity, we propose a novel method, \textit{FedQuad}, that explicitly optimises smaller intra-class variance and larger inter-class variance across clients, thereby decreasing the negative impact of model aggregation on the global model over client representations. Our approach minimises the distance between similar pairs while maximising the distance between negative pairs, effectively disentangling client data in the shared feature space. We evaluate our method on the CIFAR-10 and CIFAR-100 datasets under various data distributions and with many clients, demonstrating superior performance compared to existing approaches. Furthermore, we provide a detailed analysis of metric learning-based strategies within both supervised and federated learning paradigms, highlighting their efficacy in addressing representational learning challenges in federated settings.

\end{abstract}


\section{Introduction}
As neural network architectures have become deeper and larger over the years, their training demands increasingly large datasets. Although deep learning models thrive on vast datasets like ImageNet \cite{deng2009imagenet} or \cite{schuhmann2022laion}, such large datasets are not always available. Instead, datasets are frequently distributed across various places such as university servers, research laboratories, and private institutions. Data decentralisation presents a significant challenge for the practical applications of Deep Learning 
In addition, data privacy and protection concerns are escalating, as many entities are unwilling to share their local data due to confidentiality issues. To address these challenges, \textit{Federated Learning (FL)} has emerged as a promising solution in recent years, enabling collaborative model training without compromising data privacy \cite{mora2024enhancing, zhu2021federated}. FedAvg \cite{fedavg} is a fundamental algorithm in FL that trains a server across multiple clients. Each client preserves its local training dataset privately, ensuring that raw data is never shared with the server. Instead, clients compute model updates independently and transmit only these updates to the server, preserving data privacy while enabling collaborative learning. FL is crucial for various high-impact applications, including medical imaging, remote sensing, and image classification, where data privacy and decentralisation are paramount. 

\begin{figure} [h]
    \centering
    \includegraphics[width=0.85\linewidth]{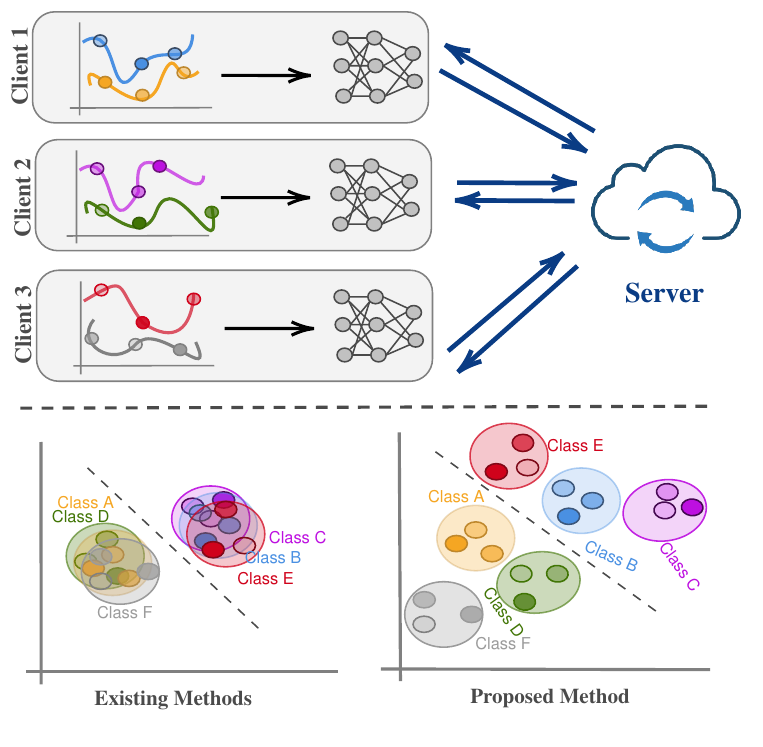}
    \caption{The representational collapse problem in FL. While clients may learn well-separated embeddings within their local classifiers, differences in data distributions across clients lead to conflicting feature spaces. After model aggregation, this discrepancy causes the global model's embeddings to collapse, leading to the loss of inter-class separability and discriminative structure.}
    \label{fig:problem}
\end{figure}

The statistical heterogeneity of local data distributions is an important challenge to federated learning. Client datasets are rarely identically distributed in practical deployments, with significant differences in class proportions, sample sizes, and feature distributions. This data heterogeneity introduces fundamental challenges for learning a global model that generalises effectively across clients. As a result, the performance of FL algorithms might suffer significantly, especially in circumstances with limited data or a high class imbalance, which are common in real-world applications.

Representational collapse occurs when feature space lacks discriminative ability, causing embeddings from different classes or clients to group identically. Figure \ref{fig:problem} illustrates the effect of data heterogeneity in federated learning. When there is data heterogeneity among client data distributions, clients are trained on varying local distributions, leading to inconsistent and incompatible feature representations. When these misaligned representations combine on the server, the global model collapses and cannot maintain significant distinctions among classes, resulting in overlapping or naive embeddings that hinder generalisation.

There are several studies to tackle data heterogeneity problems in the local training phase. FedProx  \cite{fedprox} adds the Euclidean distance between the global and server model parameters to the objective function to ensure that local updates do not deviate too much from the global model, making the updates smoother and more stable. SCAFFOLD \cite{karimireddy2019scaffold} presents a stochastic algorithm to solve the client drift problem by gradient dissimilarity using control variables. These approaches demonstrate that leveraging global model knowledge enhances the robustness of local model representations.

Beyond regularisation-based approaches, contrastive learning has emerged as an effective technique for leveraging global knowledge in FL. MOON \cite{li2021model} tackles the data heterogeneity problem by maximising the agreement between local and global model representations, demonstrating how global knowledge enhances local model robustness. FedRCL \cite{seo2024relaxed} relies on a similar perspective to the MOON methodology by ensuring that data points from the same class are sufficiently well-separated in the feature space. This separation preserves diversity, enabling the model to learn more effectively and generalise better.

While aligning local and global representations is essential for effective federated learning, overly aggressive alignment can inadvertently cause representational collapse, where intra-class diversity is diminished within client models. This phenomenon hinders the model’s ability to distinguish between minor variations within the same class, finally degrading both local and global performance. Overcoming such collapse requires a careful balance between intra-class variance and inter-class variance, ensuring that learned representations remain both discriminative and diverse across clients. This leads to a critical open question:
\textit{“How can we preserve both intra-class and inter-class relations to prevent representational collapse in global and local models under data heterogeneity?”}

We propose FedQuad, a metric learning-based federated learning approach that introduces a novel loss function to mitigate representational collapse under data heterogeneity. Unlike traditional contrastive or triplet-based methods \cite{khosla2020supervised}, \cite{hoffer2015deep}, FedQuad explicitly models the relative distances between samples by constructing stochastic quadruplets: an anchor, a positive sample (same class), a negative sample (different class), and a harder negative sample (also from a different class). The loss function simultaneously minimises the distance between the anchor and the positive while maximising the distance between the anchor and both negatives. This formulation encourages the model to learn a feature space in which intra-class samples are tightly clustered, while inter-class samples remain well-separated. By preserving both intra-class diversity and inter-class discriminability, FedQuad provides robust representations that are resilient to aggregation-induced degradation, directly addressing the core challenges of representational collapse in federated learning.
To summarise, our major contributions are as follows:
\begin{itemize}
\item We propose a novel metric learning-based federated learning framework designed to address representation collapse under data heterogeneity.
\item We analyse the impact of intra-class variance and inter-class variance, and their roles in preserving discriminative representations across clients.
\item We introduce a novel quadruplet-based loss function that effectively mitigates representational collapse in both local and global models.
\item We design an offline stochastic quadruplet sampling strategy per client, tailored to imbalanced and non-i.i.d. data distributions, to ensure robust and diverse training.
\end{itemize}
Traditional quadruplet loss \cite{chen2017beyond} primarily focuses on minimising the distance between the anchor and the positive sample, while providing only a weak push between the anchor and negative samples. Typically, it enforces a margin between a single negative pair, offering limited guidance on how to handle multiple or harder negatives. 
As a result, its effectiveness diminishes in highly heterogeneous settings such as FL, where negative samples can vary widely in difficulty. Without explicitly modelling or emphasising strong separation from challenging negatives, the learned representations may lack sufficient discriminative structure, reducing their utility in preserving inter-class boundaries.

Hard negative mining is an important step in contrastive and triplet loss-based algorithms \cite{xuan2020hard}. In our method, we present a class-separation-aware method to generate quadruplet batches, ensuring that each batch includes at least one correct positive pair and that each of the negative samples is from a different class than the anchor. This construction paradigm enables more relevant and informative optimisation, especially in federated contexts where client data distributions might vary substantially. In addition, our approach outperforms typical contrastive losses in under-represented (rare) classes, where there are generally insufficient informative negatives.

The rest of the paper is organised as follows. We review the existing works in Section \ref{sec:2}, and Section \ref{sec:3} introduces the methodology in detail. Section  \ref{sec:4} and \ref{sec:5} present empirical evaluations that demonstrates the effectiveness of the method. Finally, Section \ref{sec:6} discusses the paper and outlines potential directions for future work.

\section{Background}
\label{sec:2}

\subsection{Federated Learning}
FL frameworks such as FedAvg typically involve three main steps: \textit{broadcasting}, \textit{local training}, and \textit{model aggregation}. After each communication round, the central server broadcasts the current global model to all participating clients. Each client then performs local training on its private data, ensuring data privacy by keeping raw data on-device. Once local updates are completed (e.g., after a fixed number of epochs), clients send their updated model parameters back to the server. The server performs model aggregation, typically by averaging the received parameters, to form a new global model, which is then broadcast in the next round.
Following the typical FL approach, the data $D$ is distributed across $K$ clients. Let $D_k$ be the local dataset at client $k$, with $n_k = |D_k|$ denoting the number of data points at client $k$. The global objective function in FL is then a weighted average of the local objectives
\begin{equation}
\min _w F(w)=\sum_{k=1}^K \frac{n_k}{n} F_k(w)
\end{equation}
where $n = \sum_{k=1}^{K} n_k$ is the total number of data points across all clients and $F_k(w)$ is the local objective function at a client $k$
\begin{equation}
 \quad F_k(w)=\frac{1}{n_k} \sum_{i \in \mathcal{D}_k} f_i(w)
\end{equation}
To the best of our knowledge, this is the first work to explore the integration and analysis of metric learning losses for representation learning under federated learning settings, particularly during local client updates in the presence of data heterogeneity.
Previous research has mostly focused on two complementary ways to address heterogeneity in federated learning: improving local training and optimising global aggregation. This work belongs to the local training improvement.\\

\textbf{Federated Metric Learning} \\
Existing federated metric learning methods, such as FedMetric \cite{park2021federated}, learn embeddings using a metric loss that treats positive and negative pairs differently. However, they often fail to explicitly model the relative similarity between a given positive sample and multiple negative samples, limiting their ability to separate fine-grained structures in the representation space. Additionally, this method relies on proxy-based hypersphere clustering, which oversimplifies the underlying data distribution. Other methods, such as \cite{tian2022privacy}, \cite{gu2023defending}, and \cite{shao2023privacy}, focus on designing or enhancing the overall metric learning models in a federated setting, rather than explicitly applying metric learning losses (e.g., contrastive, triplet, or quadruplet loss) during local training on each client.

While metric learning has proven effective in supervised representation learning by minimising intra-class variance and maximising inter-class variance, including Quadruplet loss \cite{chen2017beyond}, Triplet loss \cite{hoffer2015deep}, and supervised contrastive loss \cite{khosla2020supervised}, its integration into FL remains underexplored.  Our work is among the first to systematically investigate metric learning losses in the federated setting, focusing on their role in preventing representational collapse under severe data heterogeneity. \\

\textbf{Data heterogeneity in Federated Learning} \\ 
A central challenge in federated learning (FL) is the inherent non-i.i.d. nature of data distributed across clients. In practice, clients often exhibit significant class imbalance or domain-specific biases in their local datasets, which can substantially hinder the convergence speed and generalisation performance of the global model. To solve the challenge, several techniques have been presented \cite{gao2022feddc}, \cite{feddyn}, \cite{fang2025robust}.
Contrastive learning-based approaches have gained considerable attention in federated learning (FL) for their effectiveness in aligning global and local representations, thereby mitigating the impact of client drift and data heterogeneity. For instance, FedProc\cite{mu2023fedproc}, FedCRL \cite{huang2024fedcrl} and FedPCL \cite{tan2022federated} reformulate contrastive objectives to reduce the divergence between local and global models, thereby improving alignment. Similarly, relaxed contrastive approaches such as MOON \cite{li2021model}, FedRCL \cite{seo2024relaxed}, and FedTrip \cite{li2023fedtrip} employ model-level alignment and distribution-aware aggregation to mitigate the negative impact of data heterogeneity. MOON introduces a model-contrastive loss, which aims to align the current local model with the global model while pushing the current model away from the local model of the previous round. 
In addition, unsupervised contrastive learning techniques like FedSimCLR \cite{louizos2024mutual} and FedMoCo \cite{dong2021federated} have been explored for federated settings, leveraging mutual information maximisation without requiring labelled data. However, these methods fall outside the scope of our work, as our focus lies in supervised image classification tasks where labelled data is available on each client.
In contrast to prior methods, our framework neither requires an additional global model for contrastive learning nor depends on global models or prototypes to address deviations in local training. \\

\section{Methodogy}
\label{sec:3}
\begin{figure*}[htp]
    \centering
    \includegraphics[width=0.9\linewidth]{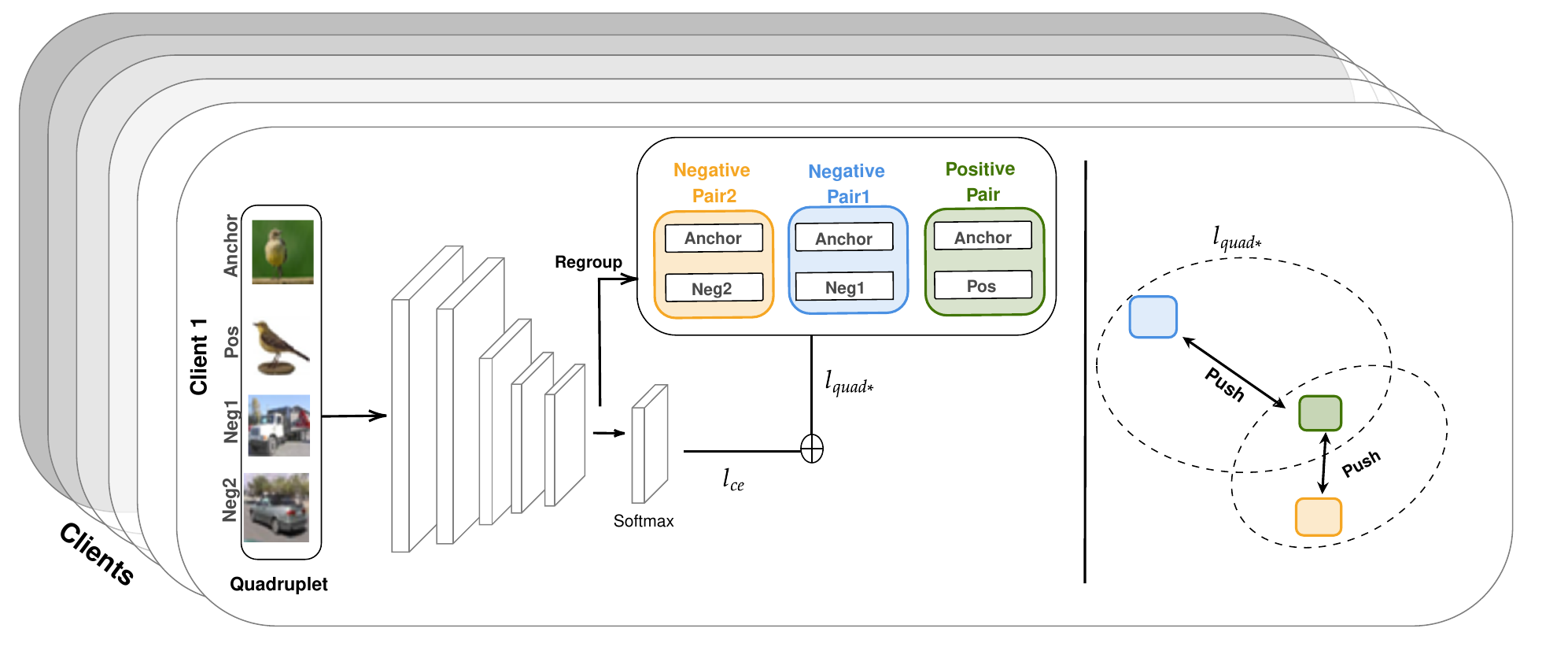}
    \caption{Overview of the FedQuad local training framework. Each client minimises loss composed of the cross-entropy loss $\ell_{ce}$, computed after the softmax layer, and the proposed quadruplet loss $\ell_{quad*}$, applied to the non-normalised embeddings from the encoder. The images are from the CIFAR-10 dataset, which explains the low resolution.}
    \label{fig:method}
\end{figure*}

Our proposed methodology is built upon the quadruplet loss framework for local training. Unlike the standard quadruplet loss, which contains a negative pair distance term ($n_1, n_2$) to limit the distance between two negative samples, we advance the focus to modelling the anchor's relation via multiple negative samples chosen from different classes. The negative pair component in traditional quadruplet loss frequently provides a weak push between negative pairs and may not contribute effectively to discriminative representation learning. Rather than explicitly modelling the structure among the negatives, we encourage the positive sample to be well-separated from the entire set of negatives collectively. 

As the negative samples originate from different classes, encouraging broader separation from the anchor, rather than focusing solely on hard negatives, can enhance the discriminative capacity of the representations. 
Our loss function is designed to capture two key objectives: the first component enforces low intra-class variance, ensuring that features from the same class remain close in the embedding space; the second component optimises the distance between classes by simultaneously pushing the anchor away from multiple negatives. The presented loss function facilitates effective feature separation both at the local client models and within the globally aggregated model.

\subsection{Stochastic Quadruplet Sampling}
We designed a stochastic pair sampling strategy to generate quadruplet samples for representation learning. Each data item consists of an anchor image, a positive image sampled from the same class as the anchor, and two negative images drawn from other classes, specifically, classes that differ both from the anchor's class and from each other. In order to enable efficient sampling, the class first constructs a mapping between class labels and sample indices inside the provided subset. During generation samples for each batch, a positive sample is chosen at random from the anchor's class, ensuring that it is not the same as the anchor itself. Negative samples are selected by first identifying all classes present in the subset, eliminating the anchor's class, then sampling two different classes and extracting one instance from each. This class-aware sampling approach ensures semantically appropriate quadruplets, allowing for the learning of discriminative and robust feature representations, especially under class imbalance or non-i.i.d. limitations in federated learning. 
$\mathcal{D}=\left\{\left(x_i, y^i\right)\right\}_{i=1}^N \text { be the dataset with inputs } x_i \in \mathbb{R}^d \text { and labels } y^i \in\{0,1, \ldots, K-1\} .$
\begin{itemize}
    \item $x^a \sim \mathcal{D}_k$
    \item $x^p \sim \mathcal{D}_k \backslash\left\{x^a\right\}$
    \item $x^{n1} \sim \mathcal{D}_{k^{\prime}} \quad$ where $k^{\prime} \in\{0, \ldots, K-1\} \backslash\{k\}$
    \item  $x^{n2} \sim \mathcal{D}_{k^{\prime \prime}} \quad$ where $k^{\prime \prime} \in\{0, \ldots, K-1\} \backslash\left\{k, k^{\prime}\right\}$
\end{itemize}

This sampling approach facilitates training with the quadruplet loss, which enforces the following constraints in the embedding space:
\begin{itemize}
    \item Positive pairs $(x^a,x^p)$ are pulled close together (minimazing intra-class variance),
    \item Anchor $x^a$ is pushed away from two distinct negatives $(x^{n1},x^{n2})$ and enhancing robustness,
    \item Selecting two distinct negatives encourages higher inter-class variance.
\end{itemize}
   
Such structured sampling is particularly beneficial in metric learning and federated learning settings, where it improves generalisation and discriminative ability under data heterogeneity.

As illustrated in Figure \ref{fig:method}, our local loss comprises two parts. The first part is a typical supervised loss (e.g., cross-entropy loss) denoted as $\ell_{ce}$. The second part is our reformulated quadruplet loss denoted as $\ell_{quad*}$. We define our loss as
\begin{equation}
\ell=\ell_{\text {ce }}\left(w_i^t ;(x^a, y^a)\right)+\beta \ell_{\text {quad* }}\left(wi^t ; (x^a, x^p, x^{n1}, x^{n2})\right)
\end{equation}
where $\beta$ is a hyperparameter to maintain the impact of
quadruplet loss. Our local objective is to minimise
\begin{equation}
\begin{aligned}
\min_{w_i^t} \; \mathbb{E}_{(x^a, y^a) \sim D^i} \Big[ 
& \; \ell_{\text{ce}}(w_i^t ; (x^a, y^a)) \\
& + \beta \, \ell_{\text{quad*}}(w_i^t ; (x^a, x^p, x^{n1}, x^{n2}))
\Big]
\end{aligned}
\end{equation}
where $m_1$ and $m_2$ are the values of margins in the two terms and $y^a$  refers to the class label of image $x^a$, $z_a$ is representation. The margins define the minimum desired difference between the distance from the anchor to the negatives. Our proposed loss is defined as;
\begin{equation}
\begin{aligned}
\ell_{\text {quad* }}= &  \left[\left\|f\left(x^a\right)-f\left(x^p\right)\right\|_2-\left\|f\left(x^a\right)-f\left(x^{n1}\right)\right\|_2+m_1\right]_+ \\
& + \left[\left\|f\left(x^a\right)-f\left(x^p\right)\right\|_2-\left\|f\left(x^a\right)-f\left(x^{n2}\right)\right\|_2+m_2\right]_+
\end{aligned}
\end{equation}
where $[x]_+ = max(0, x)$, $f(\cdot)$ is a function to extract embeddings. We choose two distinct margin values in our loss to avoid enforcing the same separation radius for both negative samples. This allows more flexibility in the embedding space, preventing all negative instances from being pushed away uniformly, and instead adapting the separation based on their semantic or class-wise dissimilarity. 

The overall federated learning algorithm is shown in Algorithm \ref{alg:alg1} that represents the FedQuad design. FedQuad remains applicable when only a small subset of clients participates in each federated learning round.  We follow the standard FedAvg approach for model aggregation; each client maintains a local model, which is synchronised with the global model regularly and updated with the local models from the clients participating in that round.
\begin{algorithm}
\caption{FedQuad Framework}
\label{alg:alg1}
\KwIn{Dataset $D$, Number of communication rounds $T$, number of clients $N$, local epochs $E$, $B$ batch size, hyperparameters $\beta$, $m_1$, $m_2$}
\KwOut{Global model $w^T$}

\textbf{Server} \\
Initialize global model $w^0$ \\
\For{$t = 0, 1, \dots, T-1$}{
    \For{$i = 1$ \KwTo $N$}{
        Get global model $w^t$ to update client $C_i$ \\
        $w_i^t \leftarrow$ \textsc{\textbf{trainClient}}$(i, w^t)$
    }
    $w^{t+1} \leftarrow \sum_{k=1}^{N} \frac{|D_i|}{|D|} w_k^t$
}
\Return $w^T$

\vspace{1em}
\textbf{Function} \textsc{\textbf{trainClient}}$(i, w^t)$:
\BlankLine
$w_i^t \leftarrow w^t$ \\
\For{$e = 1$ \KwTo $E$}{
    \For{each batch $b = \{x^a_i, x^p_i, x^{n1}_i, x^{n2}_i, y^a_i\}_{i < B}$ from $\mathcal{D}_i$}{
 
        $z_a \leftarrow f_{w_i^t}(x^a)$ \\
        $z_{\text{p}} \leftarrow f_{w_i^t}(x^p)$ \\
        $z_{\text{n1}} \leftarrow f_{w_i^t}(x^{n1})$ \\
        $z_{\text{n2}} \leftarrow f_{w_i^t}(x^{n2})$ \\
        $\ell_{\text{quad*}} \leftarrow \left[d\left(z_a, z_p\right)^2-d\left(z_a, z_{n1}\right)^2+ m_1\right]_{+} + \left[ d\left(z_a, z_p\right)^2-d\left(z_a, z_{n2}\right)^2+ m_2\right]_{+} $ \\
        $\ell_{\text{ce}} \leftarrow \text{CrossEntropyLoss}(F_{w_i^t}(x^a), y^a)$ \\
        $\ell \leftarrow \ell_{\text{ce}} + \beta \cdot \ell_{\text{quad*}}$ \\
        $w_i^t \leftarrow w_i^t - \eta \cdot \nabla \ell$
    }
}
\Return $w_i^t$
\end{algorithm}
Table \ref{tab:c10all} and \ref{tab:c100all}, our version of QuadrupletFL and FedQuad, consistently outperforms the standard supervised federated metric learning baseline. Although the modified loss function introduces twice as two negative pairs compared to the traditional triplet loss, it achieves superior performance by effectively leveraging a richer set of negative relationships. While triplet loss focuses on a single positive-negative pair at a time, incorporating multiple negative samples better facilitates the maximisation of inter-class variance, leading to more discriminative representations.

\section{Experiment}
\label{sec:4}
We compare our proposed method against supervised metric learning approaches, including triplet loss, supervised contrastive loss, and quadruplet loss. To ensure a fair comparison, we implement and evaluate these losses within a federated learning setting. Additionally, we include a baseline supervised version trained on the entire dataset. Our experiments are conducted on the CIFAR-10 and CIFAR-100 datasets \cite{krizhevsky2009learning}.
\subsection{Experimental Setup}
Our model consists of a convolutional neural network (CNN) backbone followed by a fully connected layer to produce embeddings of a specified dimension. The CNN backbone comprises three convolutional blocks. Each block includes a 2D convolutional layer with a kernel size of 3$\times$3, stride 1, and padding 1, followed by batch normalisation and a ReLU activation. The first two blocks conclude with a 2$\times$2 max pooling layer to reduce spatial dimensions. The final convolutional block uses an adaptive average pooling layer to produce a fixed-size output of 1$\times$1 spatial dimensions. The output of the convolutional backbone is flattened and passed through a fully connected layer that maps the 256-dimensional feature vector to the desired embedding dimension, which is 128 in our experiments. The forward pass applies the CNN layers, flattens the output, and then applies the linear layer to obtain the final embeddings. Additionally, a softmax layer is included solely for cross-entropy loss measurement. We make the code for our experiments publicly available to ensure reproducibility \footnote{https://anonymous.4open.science/r/FedQuad-55C8/README.md}. 
We use the Adam optimiser with a learning rate of 0.001 for all approaches. The Adam weight decay is set to $10^{-5}$, and the momentum is fixed at 0.9. The batch size is 128. For all federated learning approaches, the number of local epochs is set to 5 unless otherwise specified. The number of communication rounds is set to 20 for CIFAR-10 and CIFAR-100, as additional rounds yield little to no accuracy improvement.

Following prior works \cite{fedavg}, \cite{li2021model}, we employ a Dirichlet distribution to generate non-i.i.d. data partitions among clients, with concentration parameters $\alpha=0.5$ and $\alpha=0.3$ (lower $\alpha$ indicates highly skewed data distribution). This partitioning strategy results in some clients having relatively few or even no samples for certain classes. We evaluate scenarios with 10, 50, and 200 clients. The resulting data distributions across parties are illustrated in Figure \ref{fig:data-dist}. The best $\beta$ for reformulated quadrupled loss is 0.5, and for margin values $m_1=1.0$,  $m_2=0.5$ which is shown in an ablation study Table \ref{tab:abl}.

\section{Results}
\label{sec:5}
We compare the proposed method, FedQuad, against several metric learning-based federated learning baselines under varying clients and data distribution settings, using the CIFAR-10 and CIFAR-100 datasets. As shown in Table \ref{tab:c10all}, FedQuad consistently outperforms all baseline methods across different levels of data heterogeneity. Notably, Table \ref{tab:c10many} demonstrates that our method maintains high performance even under highly non i.i.d. distributions with many clients. Also, our method holds for the CIFAR-100 dataset, as seen in Table \ref{tab:c100many} and Table \ref{tab:c100all}.

FedQuad explicitly maximises inter-class variance within each client's representation space, which enhances feature discrimination. This is particularly beneficial when evaluating the global model on test data, which typically includes samples from all clients. Even in cases where certain clients have not encountered specific class samples, FedQuad enables the global model to learn robust representations.
\begin{table}[htbp]
\small
\centering
\renewcommand{\arraystretch}{1.2} 
\begin{tabular}{llll}
\hline
\textbf{Method} & & & \\
\hline
\textit{{\color{darkgray} Supervised (Non-FL) }} \\
Supervised  & \textbf{82.57} \\
SupCon & 76.67 \\
Triplet &  64.52 \\
Quadruplet & 67.72 \\
\hline
\textbf{Method}  & {i.i.d.} & {$\alpha=0.3$} & {$\alpha=0.5$}  \\
\hline
\textit{{\color{darkgray} Supervised FL }} \\
FedAvg   &  81.26  & 74.05  & 77.34  \\
SupConFL &  71.06 & 68.07  & 69.46 \\
TripletFL &  59.67 &  40.42 & 60.77 \\
QuadrupletFL & 62.79 & 47.27 & 64.18   \\
MOON &  76.86  & 49.69 & 61.43  \\
FedQuad  &  \textbf{82.35} & \textbf{80.13} & \textbf{80.83} \\

\hline
\end{tabular}
\caption{Accuracy ($ \% $) on CIFAR-10 across various data distributions using 10 clients, comparing supervised learning and federated learning variants.}
\label{tab:c10all}
\end{table}

\begin{table}[htbp]
\small
\centering
\renewcommand{\arraystretch}{1.2} 
\begin{tabular}{llll}
\hline
\textbf{Method} & c = 10 & c = 200\\
\hline
FedAvg & 77.34    & 60.20   \\
SupConFL &  69.46  &  36.97 \\
TripletFL& 60.77    & 55.39  \\
QuadrupletFL & 64.18  & \textbf{59.37} \\
MOON & 61.43  &  33.51 \\
FedQuad & \textbf{80.83}   &  58.89 \\

\hline
\end{tabular}
\caption{Accuracy ($\%$) on CIFAR-10 under non-i.i.d. data distribution ($\alpha = 0.5$) with many clients.}
\label{tab:c10many}
\end{table}

\begin{table}[htbp]
\small
\centering
\renewcommand{\arraystretch}{1.2} 
\begin{tabular}{llll}
\hline
\textbf{Method} &  & & \\
\hline
\textit{{\color{darkgray} Supervised (Non-FL) }} \\
Supervised &  \textbf{50.96} \\
SupCon & 37.33 \\
Triplet & 39.43   \\
Quadruplet &  40.33\\
\hline
 \textbf{Method} & {i.i.d.} & {$\alpha=0.3$} & {$\alpha=0.5$} \\
\hline
\textit{{\color{darkgray} Supervised FL }} \\
FedAvg  & \textbf{51.33} &  44.32 & 47.56   \\
SupConFL   & 34.79 &  36.72 & 36.21 \\
TripletFL  &  36.29 & 33.01  & 35.26\\
QuadrupletFL  &  39.49 & 33.92 & 35.96\\
MOON & 26.32  & 26.10 & 25.73 \\
FedQuad  & 51.27 & 48.55 & \textbf{50.64} \\

\hline
\end{tabular}
\caption{Accuracy ($ \% $) on CIFAR-100 across various data distributions using 10 clients, comparing supervised learning and federated learning variants.}
\label{tab:c100all}
\end{table}

\begin{table}[htbp]
\small
\centering
\renewcommand{\arraystretch}{1.2} 
\begin{tabular}{llll}
\hline
\textbf{Method} & c = 10 & c = 200\\
\hline
FedAvg & 47.56  & 24.51  \\
SupConFL & 36.21  & 19.44  \\
TripletFL&  35.26  & 23.79  \\
QuadrupletFL & 35.96 &  25.54  \\
MOON &  25.73 &  1.08 \\
FedQuad & \textbf{50.64} & \textbf{26.76}  \\

\hline
\end{tabular}
\caption{Accuracy ($\%$) on CIFAR-100 under non-i.i.d. data distribution ($\alpha = 0.5$) with many clients.}
\label{tab:c100many}
\end{table}
Tables \ref{tab:c100many} and \ref{tab:c10many} demonstrate that a larger number of clients can exacerbate data sparsity and increase diversity, which in turn hinders the performance of many existing methods. This effect is particularly evident on CIFAR-100, which contains 100 classes with only 500 samples per class. When data is distributed among 200 clients, the number of samples per class per client becomes very limited, making it challenging to learn robust representations.

Despite this challenge, our method effectively handles both data imbalance and sparsity under non-i.i.d. data distributions. In contrast, MOON demonstrates substantially degraded performance under these conditions, highlighting the limitations of global model regularisation–based contrastive learning approaches in scenarios with a large number of clients and highly diverse data distributions. These findings indicate that MOON struggles to generalise and fails to learn robust representations when provided with serious data heterogeneity.

\begin{figure}[ht]
  \centering
  \begin{minipage}[t]{0.48\linewidth}
    \centering
    \includegraphics[width=\linewidth]{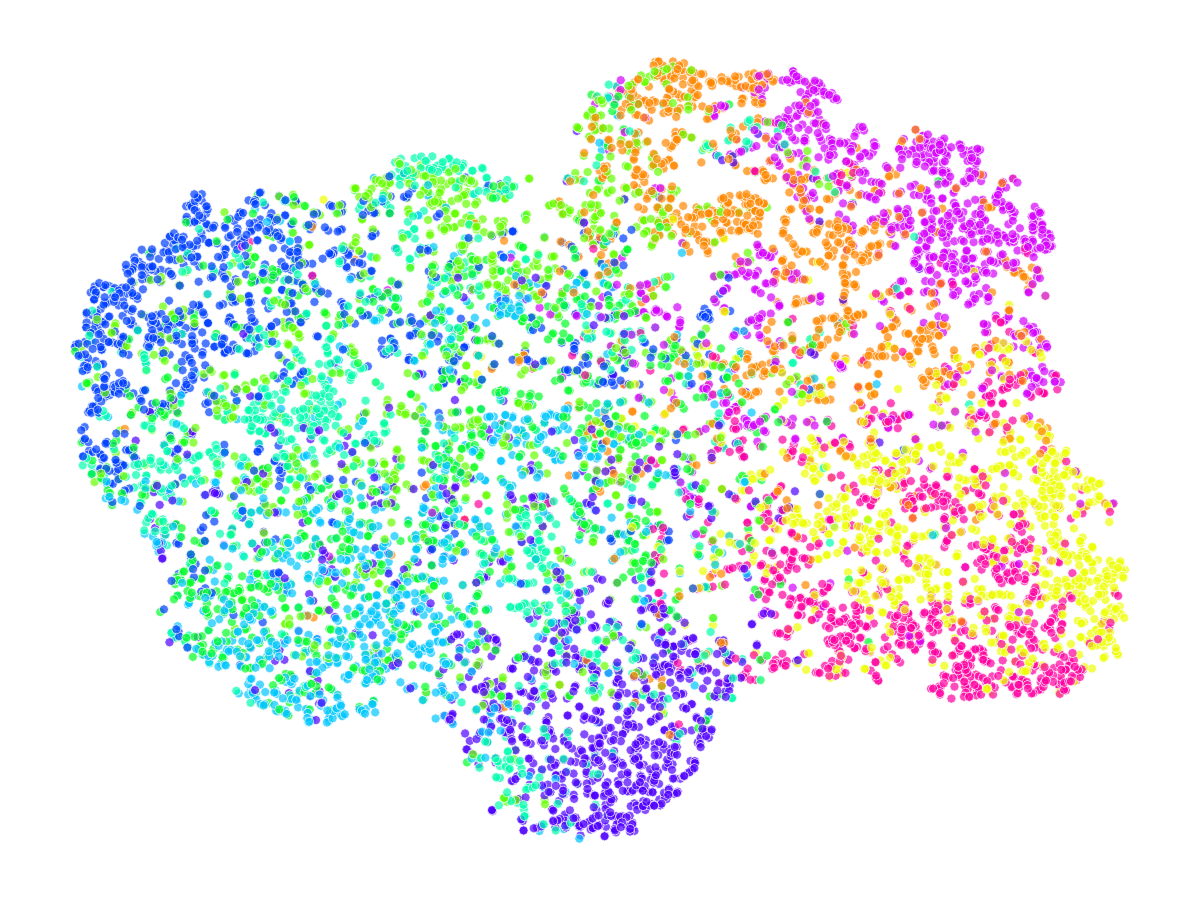}
    \caption*{(a) FedAvg}
  \end{minipage}
  \hfill
  \begin{minipage}[t]{0.48\linewidth}
    \centering
    \includegraphics[width=\linewidth]{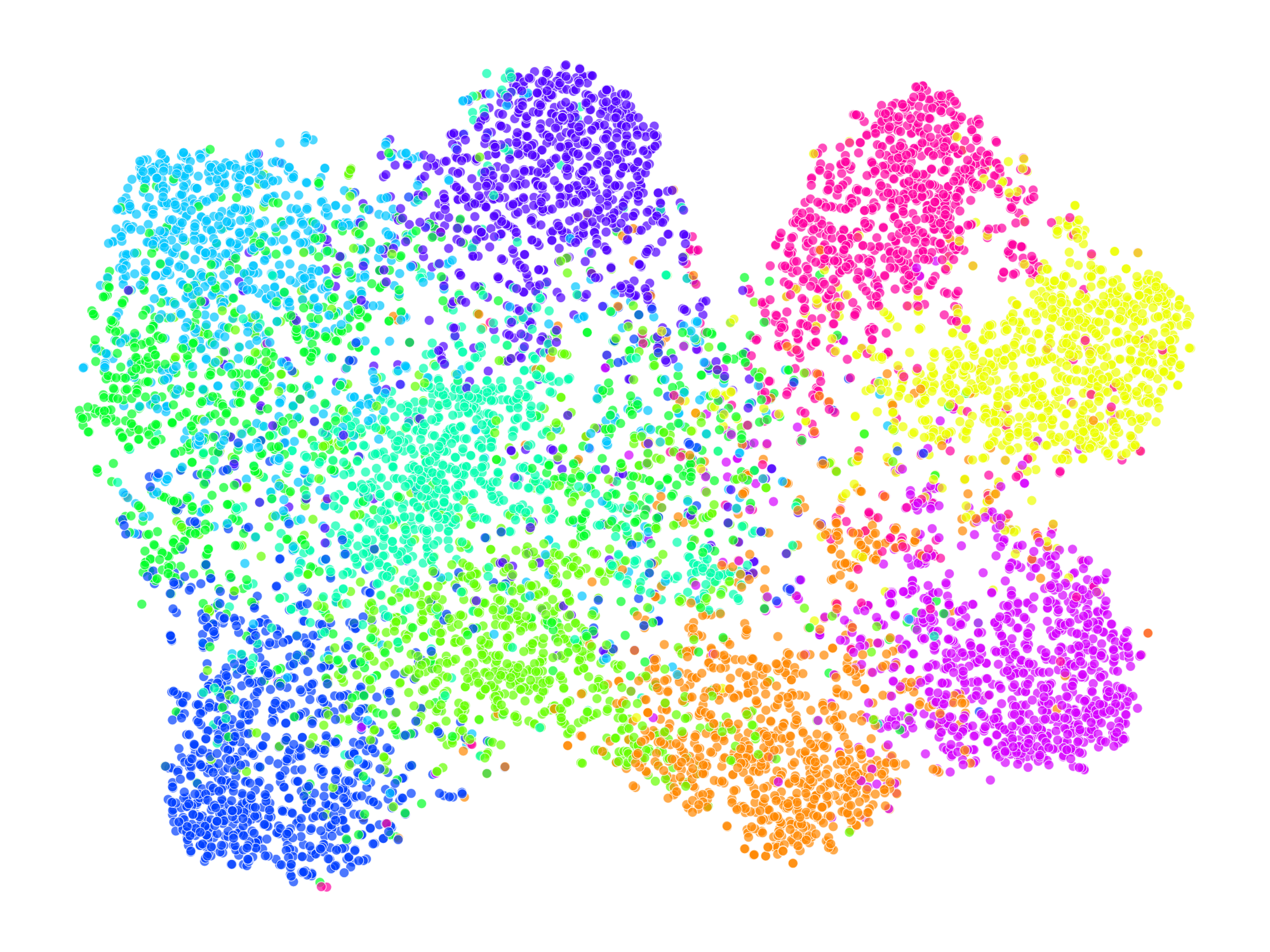}
    \caption*{(b) FedQuad}
  \end{minipage}
    \vspace{0.8em}
  \begin{tikzpicture}[scale=0.8]
    \scriptsize     

 \foreach \i/\name/\col in {
    0/airplane/1!80,
    1/automobile/2!80,
    2/bird/3!80,
    3/cat/4!80,
    4/deer/5!80,
    5/dog/6!80,
    6/frog/7!80,
    7/horse/8!80,
    8/ship/9!90,
    9/truck/10!80
} {
        \pgfmathsetmacro{\x}{mod(\i,5)*2}
        \pgfmathsetmacro{\y}{-0.25 * int(\i/5)}
        \edef\fillcolor{\col}
        \fill[\fillcolor] (\x,\y) circle (2.5pt);
        \node[anchor=west] at (\x+0.2,\y) {\name};
    }

    \draw[line width=0.1pt] (-1,-0.5) rectangle (10,0.2);
\end{tikzpicture}
  \vspace{5pt} 

  \begin{minipage}[t]{0.48\linewidth}
    \centering
    \includegraphics[width=\linewidth]{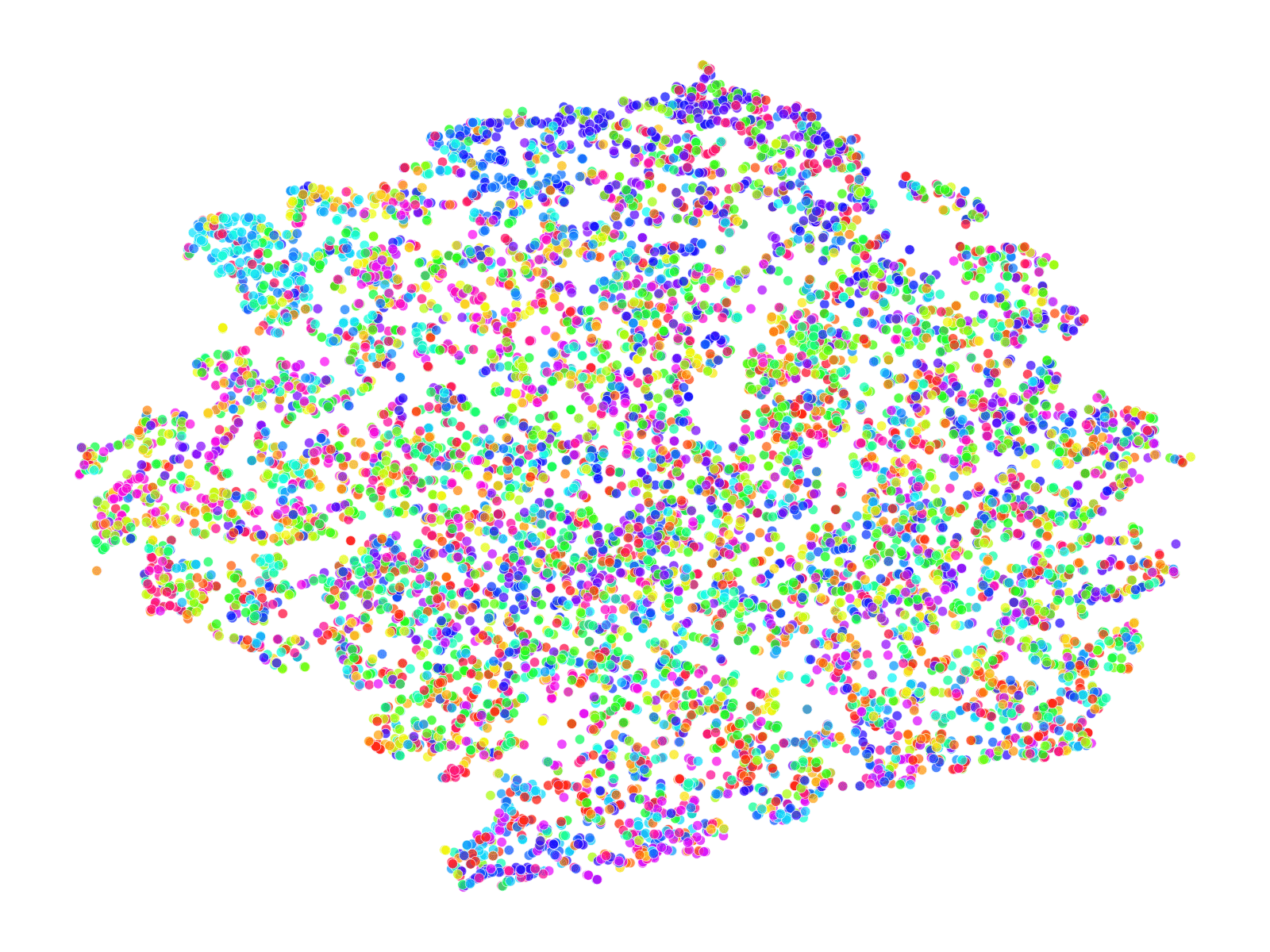}
    \caption*{(c) FedAvg}
  \end{minipage}
  \hfill
  \begin{minipage}[t]{0.48\linewidth}
    \centering
    
   \includegraphics[width=\linewidth ]{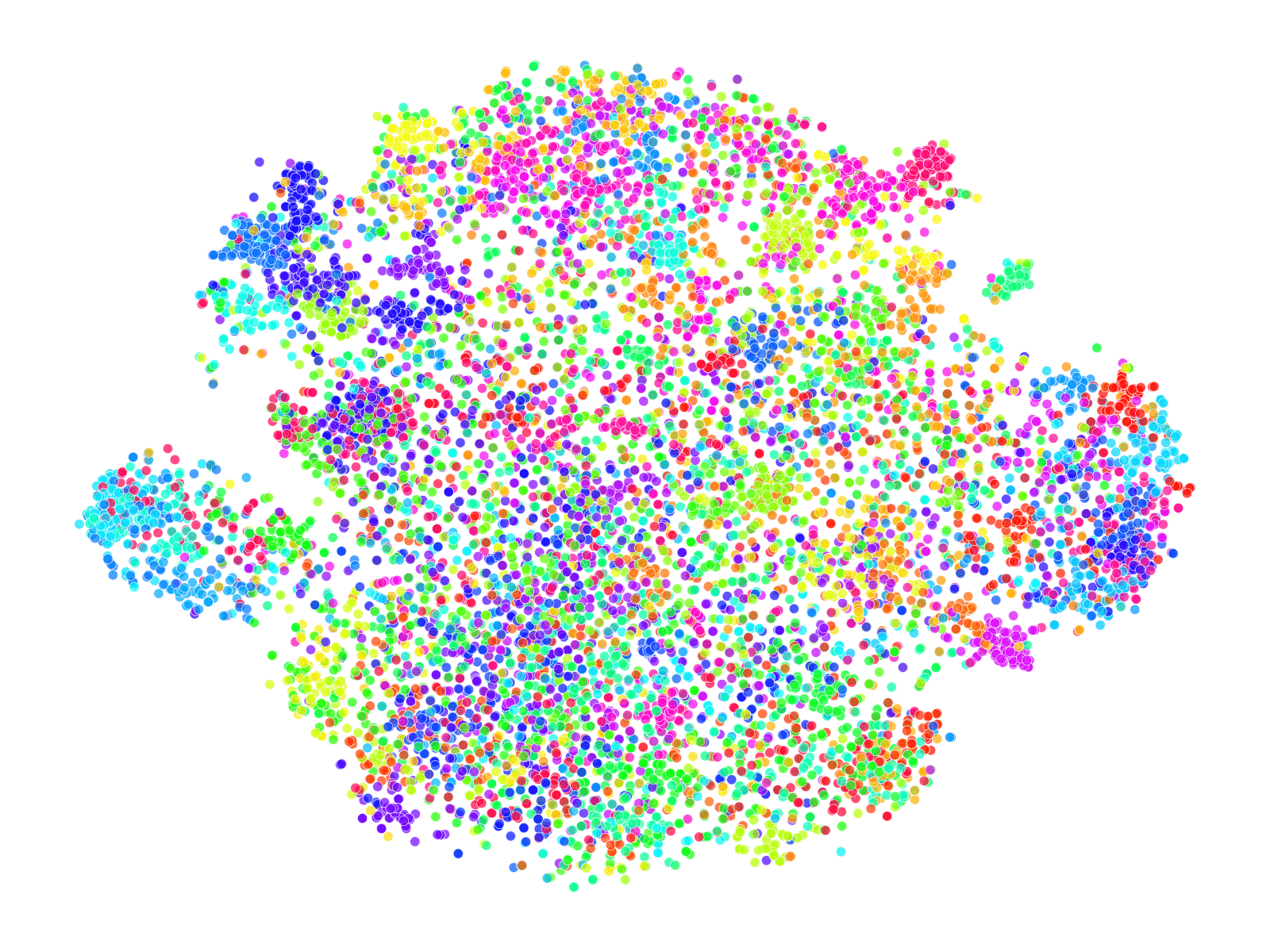}
    \caption*{(d) FedQuad}
  \end{minipage}

\begin{tikzpicture}[scale=0.8]
    \scriptsize     

    \foreach \i/\name/\col in {
        0/bee/1!80,
        1/bus/2!80,
        2/fox/3!80,
        3/forest/7!80,
        4/girl/4!80,
        5/lizard/5!80,
        6/lion/6!80,
        7/house/8!80,
        8/lobster/9!80,
        9/mountain/10!80,
        10/mouse/11!80,
        11/road/12!80,
        12/snake/13!80,
        13/tank/14!80,
        14/wolf/15!80
    } {
        \pgfmathsetmacro{\x}{mod(\i,5)*2}
        \pgfmathsetmacro{\y}{-0.25 * int(\i/5)}
        \edef\fillcolor{\col}
        \fill[\fillcolor] (\x,\y) circle (2.5pt);
        \node[anchor=west] at (\x+0.1,\y) {\name};
    }

    \draw[line width=0.1pt] (-1.,-0.7) rectangle (10,0.2);
\end{tikzpicture}

  \caption{t-SNE visualisation of learned representations at an early stage of training (Round 5) under a non-i.i.d. data distribution. The first row shows the global model's embeddings on CIFAR-10, all ten classes. The second row presents embeddings for a subset of classes from CIFAR-100.}
  \label{fig:tsne}
\end{figure}
Figure \ref{fig:tsne} presents the global model embeddings for both datasets. Despite being at an early stage of training, t-SNE plots demonstrate that metric learning–based local training can capture discriminative and robust features, enabling effective separation of samples across classes.

\begin{figure}[ht]
  \centering

  \begin{minipage}[t]{0.48\linewidth}
    \centering
    \includegraphics[width=\linewidth]{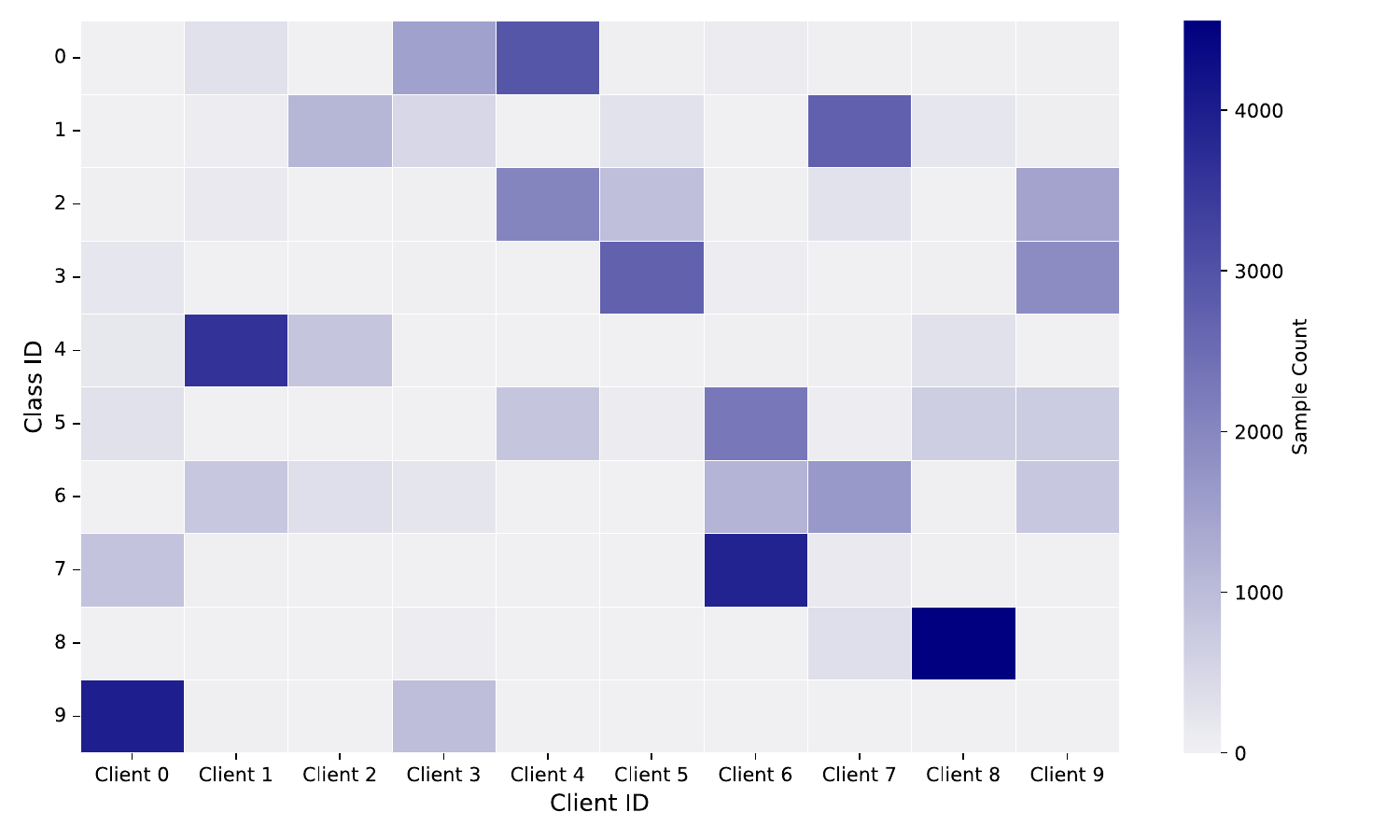}
    \caption*{(a) CIFAR-10}
  \end{minipage}
  \hfill
  \begin{minipage}[t]{0.48\linewidth}
    \centering
    \includegraphics[width=\linewidth]{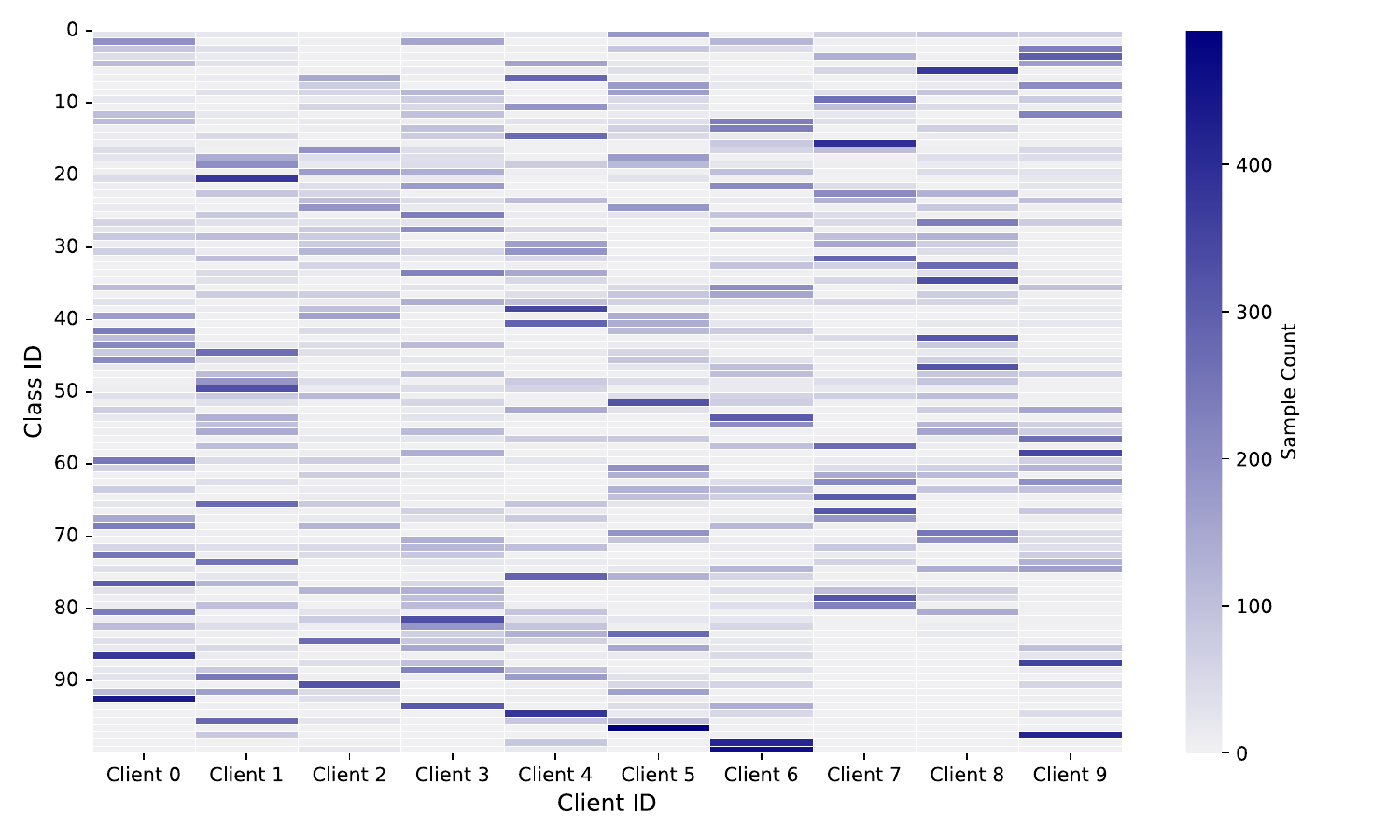}
    \caption*{(b) CIFAR-100}
  \end{minipage}

  \caption{Client-wise data distribution under non-i.i.d. partitioning. Each rectangle represents the number of samples from a specific class assigned to a particular client. The colour bar indicates the sample count, with darker shades denoting higher quantities. The visualisation highlights the heterogeneity and class imbalance introduced by the non-i.i.d. data partitioning strategy.}
  \label{fig:data-dist}
\end{figure}
Figure \ref{fig:data-dist} illustrates the number of samples per class for each client. Under highly heterogeneous settings (e.g., Dirichlet $\alpha=0.3$), we observe significant imbalance across clients, particularly on CIFAR-100, where each class has fewer samples (500 samples per class). This results in a higher degree of non-i.i.d.-ness, posing greater challenges for representation learning.

\begin{table}[htbp]
\small
\centering
\renewcommand{\arraystretch}{1.2} 
\begin{tabular}{lllll}
\hline
\textbf{Method} &$\beta$ & $m_1$ & $m_2$ & {Accuracy ($\%$)}\\
\hline
    FedQuad & 0.5 &1.0 &  1.0& 71.51 \\
    FedQuad & 0.5 & 1.0 & 0.5& \textbf{72.68}\\
    FedQuad & 1.0 & 1.0 &  0.5& 70.4 \\
    FedQuad (without $\ell_{ce} $) & 0.5 & 1.0 & 0.5& 62.41\\
    FedQuad (without $\ell_{ce} $) & 0.5 & 2.0 & 0.5 & 60.99\\
    FedQuad (without $\ell_{ce} $) & 0.5 & 5.0 & 0.5 & 57.26\\
    FedQuad (without $\ell_{ce} $) & 0.5 & 1.0 & 1.0 & 56.42\\
\hline
\end{tabular}
\caption{Ablation study on the effect of loss hyperparameters ($\beta$, $m_1$, $m_2$) in the proposed quadruplet loss, evaluated on CIFAR-10 under an i.i.d. setting with 10 clients over 5 communication rounds.}
\label{tab:abl}
\end{table}

\section{Discussion}
\label{sec:6}
Our experimental results show that FedQuad consistently enhances model generalisation in federated learning scenarios with a variety of data heterogeneity. FedQuad addresses one of the most significant challenges in FL, representational collapse, which is common under non-i.i.d. conditions, by explicitly optimising the distances between positive and negative pairs. The strategy successfully increases inter-class variance, resulting in more stable and discriminative global representations without requiring raw data exchange among clients.
Contrastive learning has shown potential in centralised settings, but its performance in federated learning is insignificant, especially under class-imbalanced distributions (e.g., Dirichlet $\alpha=0.3$). We attribute this to the fundamental constraints of contrastive loss, which frequently results in mismatched or misaligned latent representations among clients. 
In contrast, our stochastic alignment approach provides flexibility to representation learning, allowing for more effective adaptation across varied client data. This leads to consistently higher performance, particularly under extreme non-i.i.d. conditions.

Traditional contrastive or triplet-based losses tend to collapse in cases with high class imbalance, leading to noisy or overlapping feature embeddings. FedQuad successfully maintains the structure of local representations while effectively aligning them with the global feature space. This is achieved without requiring direct computation of distances between negative pairs in each batch or any data exchange, thereby maintaining strict data privacy while still aligning feature spaces in a globally coherent manner.

A key insight from our study is that achieving a balance between local discriminability and global consistency in federated learning requires deliberate loss function design. The proposed stochastic quadruplet formulation extends beyond traditional contrastive and triplet losses by introducing stronger negative constraints and finer-grained control over pairwise relations. This leads to embeddings that are significantly more robust to client-level distributional shifts. Such robustness is particularly valuable in low-resource settings or under severe class imbalance, where conventional federated learning methods often suffer from performance degradation due to poorly aligned feature spaces.

While FedQuad demonstrates strong performance under various non-i.i.d. conditions, scalability remains a potential limitation. In scenarios with many clients (e.g., 1000) and limited data per client, the number of informative negative samples decreases significantly, which may weaken the effectiveness of the stochastic quadruplet loss.

Our method focuses on positive versus two negatives simultaneously, allowing the model to push embeddings away from multiple negative directions while preserving similarity with positive pairs, a key strength that improves representation robustness. Furthermore, while our experiments on CIFAR-10 and CIFAR-100 demonstrate FedQuad's utility, extending evaluation to limited and domain-specific datasets such as medical imaging or user behaviour data would provide stronger evidence of the method's practical applicability and generalisability in real-world federated learning scenarios.

\section{Conclusion}
In this work, we introduced FedQuad, a federated learning framework based on metric learning, designed to directly address representational collapse caused by data heterogeneity among clients. Leveraging a stochastic quadruplet loss, FedQuad promotes lower intra-class variance and higher inter-class variance inside local feature spaces, thereby enhancing global representations without requiring access to raw client data. Furthermore, we provide an in-depth analysis of metric learning in federated settings, particularly under conditions where data is imbalanced and limited.

Comprehensive experiments on CIFAR-10 and CIFAR-100 across a range of non-i.i.d. scenarios demonstrate that FedQuad consistently outperforms existing baselines, especially in the presence of class imbalance and a large number of clients. These results underscore the promise of metric learning for local representation alignment and highlight the importance of structured embedding objectives in mitigating the effects of client divergence. This work establishes the way for future research, such as extending FedQuad to semi-supervised and unsupervised federated learning, enhancing hard negative mining strategies, and exploring local training objectives for robust representation alignment under heterogeneous distributions.\\
\textbf{Acknowledgements:} \"Ozg\"u G\"oksu is supported by the Turkish Ministry of National Education.

\end{document}